\documentclass[journal]{IEEEtran}
\ifCLASSINFOpdf
\else
\fi
\usepackage{mathtools}
\usepackage{color}
\usepackage{algorithm}
\usepackage[noend]{algpseudocode}

\usepackage{times}
\usepackage{epsfig}
\usepackage{graphicx}
\usepackage{amssymb}
\usepackage{romannum}
\usepackage{color}

\usepackage{amsmath}
\usepackage{breqn}
\usepackage{multirow}
\usepackage{subfigure}
\usepackage{xurl}
\usepackage{caption}

\begin{document}
%
\title{Comprehensive Validation on Reweighting Samples for Bias Mitigation via AIF360}
%
%
%

 \author{Christina Hastings Blow,~Lijun Qian,~ Camille Gibson, ~Pamela Obiomon,~Xishuang Dong
\thanks{C. Hastings, L. Qian, P. Obiomon and X. Dong are with the Department of Electrical and Computer Engineering, Prairie View A\&M University, Texas A\&M University System, Prairie View, TX 77446, USA. C. Gibson is with the College of Juvenile Justice,
Executive Director of Texas Juvenile Crime Prevention Center, Prairie View A\&M University, Texas A\&M University System, Prairie View, TX 77446, USA. Email: chastings@pvamu.edu, liqian@pvamu.edu, cbgibson@pvamu.edu, phobiomon@pvamu.edu, xidong@pvamu.edu}
}

\maketitle

\begin{abstract}
Fairness AI aims to detect and alleviate bias across the entire AI development life cycle, encompassing data curation, modeling, evaluation, and deployment—a pivotal aspect of ethical AI implementation. Addressing data bias, particularly concerning sensitive attributes like gender and race, reweighting samples proves efficient for fairness AI. This paper contributes a systematic examination of reweighting samples for traditional machine learning (ML) models, employing five models for binary classification on the Adult income and COMPUS datasets with various protected attributes. The study evaluates prediction results using five fairness metrics, uncovering the nuanced and model-specific nature of reweighting sample effectiveness in achieving fairness in traditional ML models,  as well as revealing the complexity of bias dynamics.

\end{abstract}


%
\IEEEpeerreviewmaketitle

\section{Introduction }
\label{sec1}


Artificial intelligence (AI) ethics represents an evolving and interdisciplinary domain dedicated to grappling with ethical considerations in AI~\cite{allen2006machine, anderson2007machine}. This field explores ethical theories, guidelines, policies, principles, rules, and regulations pertinent to AI~\cite{siau2020artificial}. Establishing ethical standards for AI is crucial for developing morally sound AI systems or ensuring ethical behavior in AI applications. It involves discerning ethical and moral values, guiding principles that define what is morally right or wrong. Numerous ethical issues have been identified in AI applications and studies, including concerns about transparency, privacy, accountability, bias, discrimination, safety, security, and the potential for criminal or malicious use. In recent years, AI ethics has burgeoned into a broad and rapidly advancing research area, drawing increased attention from researchers~\cite{huang2022overview}.

Fairness AI endeavors to identify and mitigate bias throughout the entire life cycle of AI technique development, spanning data curation and preparation, modeling, evaluation, and deployment, which is crucial for the successful implementation of AI ethics~\cite{mehrabi2021survey, akter2021algorithmic, ferrara2023fairness, chen2023unmasking}. Bias can manifest in various forms, potentially leading to unfairness in different downstream learning tasks. These biases typically originate from different stages of the machine learning pipeline, including data curation, algorithm design, and user interactions. Data bias may arise when the data are collected from skewed sources, are incomplete, lack crucial information, or contain errors. Such biases result in unrepresentative or incomplete data, leading to biased outputs. Algorithmic bias stems from biased assumptions or criteria in algorithm design, resulting in biased outputs for downstream tasks. Bias introduced through user interactions occurs when individuals using AI systems inject their own biases or prejudices, whether consciously or unconsciously. To address these sources of bias, various approaches have been proposed. Dataset augmentation involves adding more diverse data to training datasets, enhancing representativeness and reducing bias~\cite{sharma2020data}. Bias-aware algorithms are designed to consider different types of bias, working to minimize their impact on system outputs~\cite{ahsen2019algorithmic}. User feedback mechanisms, such as human-in-the-loop systems~\cite{ghai2022d}, involve soliciting feedback from users to identify and rectify biases in the system.

Addressing data bias, especially concerning sensitive attributes like gender and race, can be efficiently achieved through sample reweighting, contributing to the advancement of fairness in AI~\cite{kamiran2012data}. This method involves assigning weights to each sample based on the ratio of its population proportion to its sampling proportion~\cite{an2020resampling}. The process ensures the dataset becomes discrimination-free through two key steps. Firstly, specific attributes, such as gender and race, are identified within the datasets. Subsequently, higher weights are assigned to samples from underrepresented groups, while lower weights are assigned to those from overrepresented groups with respect to these specific attributes. These steps collectively contribute to achieving balance across all groups, thereby fostering fairness in the final outputs of AI algorithms trained on the reweighed data. Nevertheless, existing efforts appear to lack a thorough and comprehensive evaluation of the effectiveness of reweighting samples in mitigating bias associated with traditional machine learning models.

This paper conducts a thorough evaluation of the efficacy of reweighting samples in addressing bias associated with traditional machine learning models using AIF360~\cite{bellamy2018ai}. AI Fairness 360 (AIF360) is a versatile open-source library designed to identify and alleviate bias in machine learning models across the entire AI application lifecycle. It encompasses a comprehensive array of metrics for scrutinizing biases in datasets and models, along with detailed explanations for these metrics and algorithms for bias mitigation. The evaluation in this paper focuses on the reweighting sample methods available in AIF360, applied to classification tasks performed by five traditional machine learning models: Decision Tree, K Nearest Neighbor, Gaussian Naïve Bayes, Logistic Regression, and Random Forest. The reweighting process is carried out with respect to privileged attributes such as sex and race. Subsequently, each traditional machine learning model undergoes classification tasks on both the original datasets and the new datasets resulting from reweighting samples. The comparative analysis involves assessing the performance of these models on the original and new datasets, respectively. This evaluation is based on metrics such as balanced accuracy and fairness metrics.  Experimental results highlights the model-specific nature of reweighting sample effectiveness in achieving fairness in traditional ML models, as well as reveal the complexity of bias dynamics.

The contributions in this paper can be summarized as:

\begin{enumerate}
\item In contrast to the previous work~\cite{hufthammer2020bias}, our research involves a systematic comparison of reweighting samples for mitigating bias on five traditional ML models through AIF360 platform.

\item We systematically examine the fairness of experimental results with five fairness metrics and provide insights of effectiveness of reweighting samples for bias mitigation.
\end{enumerate}

\section{Methodology}
\label{sec2}

This paper aims to examine the effectiveness of reweighting sample to enhancing fairness of traditional machine learning algorithms on classification tasks, which covers three AI techniques, namely, reweighting samples, traditional machine learning, and fairness metrics to performance evaluation.

\subsection{Reweighting samples}

Fairness in AI can be conceptualized as a multi-objective optimization challenge, aiming to optimize learning objectives while mitigating discrimination with respect to sensitive attributes~\cite{kamiran2012data}. In essence, achieving fairness may involve a trade-off in learning performance to minimize bias. Data preprocessing emerges as an effective technique for molding training data to foster fairness in AI. Techniques such as suppression, dataset massaging, and reweighting samples have proven effective~\cite{kamiran2012data}.

Reweighting samples, a specific preprocessing technique, involves adjusting the significance or contribution of individual samples within the training dataset. By strategically assigning weights, it becomes possible to render the training dataset free from discrimination concerning sensitive attributes, all without altering the existing labels. One approach to determining these weights involves measuring them based on the frequency counts associated with the sensitive attribute~\cite{calders2009building}.

This paper leveraged the reweighting sample technique from AIF360 during data preprocessing to enhance fairness in binary classification. The input for the reweighting process comprises a training dataset with samples containing attributes (including a sensitive attribute) and labels, along with the specification of the sensitive attribute. The output is a transformed dataset where sample weights are adjusted concerning the sensitive attribute, mitigating potential classification bias. Throughout the reweighting process, an analysis of the distribution of the sensitive attribute within different groups is conducted. This analysis informs the calculation of reweighting coefficients, which, in turn, adjust sample weights to promote a more uniform distribution across groups.

Given a sensitive (protected) attribute, the privileged group of samples includes the samples with the positive sensitive attribute while  the unprivileged group of samples includes the samples with the negative sensitive attribute. For a binary classification task, reweighting samples adjusts weights of four categories of samples, namely  $w_{pp}$ (the weight of the positive privileged sample (pp)), $w_{pup}$ (the weight of positive unprivileged samples (pup)), $w_{np}$ (the weight of negative privileged samples (np)), and $w_{nup}$ (the weight of negative unprivileged samples (nup))  as below.

\begin{equation}
w_{pp} = \frac{N_{p}}{N_{total}} \times \frac{N_{pos}}{N_{pp}} 
\end{equation}

\begin{equation}
w_{pup} = \frac{N_{up}}{N_{total}} \times \frac{N_{pos}}{N_{pup}} 
\end{equation}

\begin{equation}
w_{np} = \frac{N_{p}}{N_{total}} \times \frac{N_{neg}}{N_{np}} 
\end{equation}

\begin{equation}
w_{nup} = \frac{N_{up}}{N_{total}} \times \frac{N_{neg}}{N_{up}} 
\end{equation}

where 

$N_{p}$: the number of samples in the privileged group.

$N_{pp}$:  the number of samples with the positive class in the privileged group. 

$N_{np}$:  the number of samples with the negative class in the privileged group. 

$N_{up}$:   the number of samples in the unprivileged group.

$N_{pup}$:  the number of samples with the positive class in the unprivileged group. 

$N_{nup}$: the number of samples with the negative class in the unprivileged group. 

$N_{pos}$: the number of samples with the positive class.

$N_{neg}$:  the number of samples with the negative class.

$N_{total}$: the number of samples.

\subsection{Traditional Machine Learning}

This paper applied five traditional machine learning models to implement binary classification tasks, including Logistic Regression, Decision Tree, K Nearest Neighbor, Gaussian Naive Bayes, and  Random Forest~\cite{bishop2006pattern}. 

\textbf{Logistic Regression}  is a statistical technique employed to model the probability of a binary outcome. It finds widespread application in machine learning for scenarios involving binary classification, aiming to predict whether an instance belongs to one of two classes. The posterior probability of class $c_{1}$ is expressed through a logistic sigmoid applied to a linear function of the feature vector $\phi$. 

\begin{equation}
p(c_{1}|\phi) = y(\phi) = \sigma(w^{T}\phi)
\end{equation}

while $p(c_{2}|\phi) = 1 - p(c_{1}|\phi)$ and $\sigma(\cdot)$ is the logistic sigmoid function.

\textbf{Decision Tree} is a data mining technique employed to establish classification systems using multiple covariates or to create prediction algorithms for a target variable. This method organizes a population into branch-like segments, forming an inverted tree structure with a root node, internal nodes, and leaf nodes. Notably, the algorithm is non-parametric, allowing it to effectively handle large and complex datasets without imposing a rigid parametric structure~\cite{song2015decision}. However, one limitation of decision trees lies in their reliance on hard splits in the input space, where a single model is responsible for predictions for any given value of the input variables.

\textbf{K Nearest Neighbor} is a supervised machine learning algorithm used for classification and regression tasks. It is a type of instance-based learning, where the model makes predictions based on the similarity of new data points to existing labeled data points in the training set~\cite{peterson2009k, kramer2013k}.

\textbf{Gaussian Naive Bayes}  is a straightforward classification algorithm~\cite{bishop2006pattern}. Its primary principle involves assigning labels to classes by maximizing the posterior probability for each sample. This method operates under the assumption that voxel contributions are conditionally independent and follow a Gaussian (normal) distribution. The discriminant function is formulated as the sum of squared distances to the centroid of each class across all voxels in the searchlight. This sum is then weighted by the variance and the logarithm of the a-priori probability, computed in the training set using Bayes rule. In essence, Gaussian Naive Bayes provides a probabilistic approach to classification, leveraging assumptions about the distribution of features to make predictions.

\textbf{Random Forest} has proven to be remarkably successful as a classification and regression method. This approach involves the combination of multiple randomized decision trees, and it aggregates their predictions through averaging. Notably, it has demonstrated exceptional performance in scenarios where the number of variables is significantly greater than the number of observations. Its versatility extends to large-scale problems, making it adaptable to various ad hoc learning tasks. Additionally, it provides measures of variable importance, adding to its utility and interpretability~\cite{biau2016random}.

\subsection{Fairness Metrics}

This paper employed five fairness metrics to evaluate the effectiveness of reweighting samples for mitigating bias.

Given sensitive (protected) attributes, \textbf{Disparate Impact (DI)} denotes inadvertent bias that may arise when predictions exhibit varying error rates or outcomes across demographic groups, where the sensitive attributes, such as race, sex, disability, and age, are deemed protected. This bias can emerge either from training models on biased data or from the predictive model itself being discriminatory. In the context of this study, Disparate Impact pertains to divergent impacts on prediction results as defined by.

\begin{equation}
DI =  \frac{p_{pup}}{p_{pp}} 
\end{equation}

where $p_{pup}$ refers to the prediction probability for the unprivileged samples with positive predictions  while $p_{pp}$ denotes the prediction probability for the privileged samples with positive predictions. If the disparate impact of the predictions approaches 0, it signifies bias in favor of the privileged group. Conversely, if it exceeds 1, it indicates a bias in favor of the unprivileged group. A value of 1 implies perfect fairness in the predictions~\cite{feldman2015certifying}.

\textbf{Average odds difference (AOD)} is the average of difference in the false positives rates (FPR) and true positive rates (TPR) between the unprivileged and privileged groups. It is defined by

\begin{equation}
AOD =  \frac{(FPR_{up} - FPR_{p}) + (TPR_{up} - TPR_{p}) }{2} 
\end{equation}

where  $FPR_{up}$ and $FPR_{p}$ denote the False Positive Rate for unprivileged and privileged samples, respectively, within predictions, while $TPR_{up}$ and $TPR_{p}$ refer to the True Positive Rate for unprivileged and privileged samples, respectively, within predictions. A result of 0 signifies perfect fairness. A positive value indicates bias in favor of the unprivileged group, while a negative value indicates bias in favor of the privileged group.

\textbf{Statistical parity difference (SPD)} is to calculate the difference between the ratio of favorable outcomes in unprivileged and privileged groups. It is defined by

\begin{equation}
SPD =  p_{pup} - p_{pp}
\end{equation}

A score below 0 suggests benefits for the unprivileged group, while a score above 0 implies benefits for the privileged group. A score of 0 indicates that both groups receive equal benefits.

\textbf{Equal opportunity difference (EOD)} involves evaluating the equal opportunity for benefiting all groups. EOD specifically centers on the True Positive Rate (TPR), representing the accurate identification of positives in both the unprivileged and privileged groups. The measure is defined by

\begin{equation}
EOD =  TPR_{up} - TPR_{p}
\end{equation}

A value of 0 signifies perfect fairness. A positive value indicates bias in favor of the unprivileged group, while a negative value indicates bias in favor of the privileged group.

\textbf{Theil index (TI)} is also called the entropy index which measures both the group and individual fairness. It is defined by

\begin{equation}
TI =  \frac{1}{n}\sum_{i = 1}^{n}{ \frac{b_{i}}{\mu}ln\frac{b_{i}}{\mu}}
\end{equation}

where $b_{i} = \hat{y_{i}} - y_{i} + 1$ and $\mu$ is the average of $b_{i}$. A lower absolute value of TI value in this context would indicate a more equitable distribution of classification outcomes, while a higher absolute value suggests greater disparity.

\section{Experiment }
\label{sec4}

\subsection{Dataset}

This paper employed two datasets, including the Adult Income dataset and the Correctional Offender Management Profiling for Alternative Sanctions (COMPAS) dataset, to evaluate the effectiveness of reweighting samples for mitigating fairness.

\textbf{ Adult Income dataset: } it includes $48,842$ samples with 14 attributes, which can be used for predicting whether income exceeds $50K/yr$ based on census data~\cite{adult}.  

\textbf{COMPAS dataset: } the COMPAS system is a case management system and decision support tool initially developed and owned by Northpointe (now Equivant). It was designed for the purpose of assessing the likelihood of an individual committing a future crime. The dataset associated with COMPAS comprises more than 20 attributes and includes a substantial sample size of 11,000 instances~\cite{compas}.

\subsection{Experimental metrics}

Balance Accuracy (BA) is applicable to both binary and multi-class classification scenarios by computing the mean of sensitivity and specificity. Sensitivity gauges the correct prediction of true positives, representing accurately identified positive instances, while specificity measures the true negatives over the total negatives predicted by the model. A result nearing 0 signifies poor model performance, whereas a result approaching 1 indicates effective performance across both sensitivity and specificity~\cite{brodersen2010balanced}.

\newcommand{\photo}[1]
{
    \includegraphics[width=4cm]{#1}
}

\subsection{Experimental results and discussion}

This paper implemented comprehensive validation on reweighting samples for mitigating bias for traditional supervised machine learning models with the binary classification tasks. It involved two components of experiments on two datasets: Adult income dataset and COMPAS dataset.

\subsubsection{Adult income}

\begin{figure*}[h!]
	\center
	\includegraphics[width=1.0\linewidth]{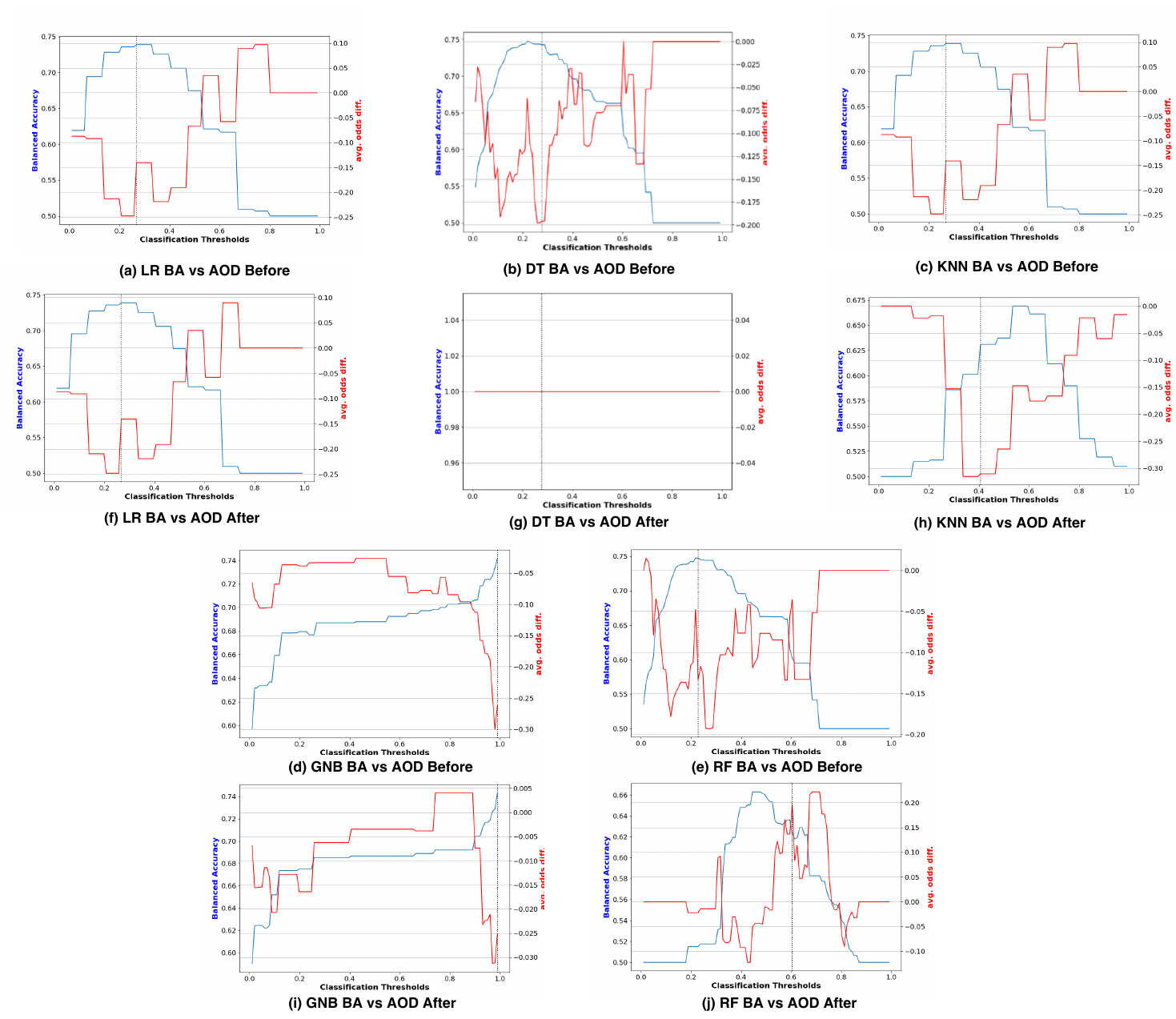}
	\caption{Performance comparison via BA vs AOD before and after reweighting samples on Adult income dataset with respect to the protected attribute $Race$.}
	\label{fig:Adult_Race_AOD}
\end{figure*}

\begin{figure*}[h!]
	\center
	\includegraphics[width=1.0\linewidth]{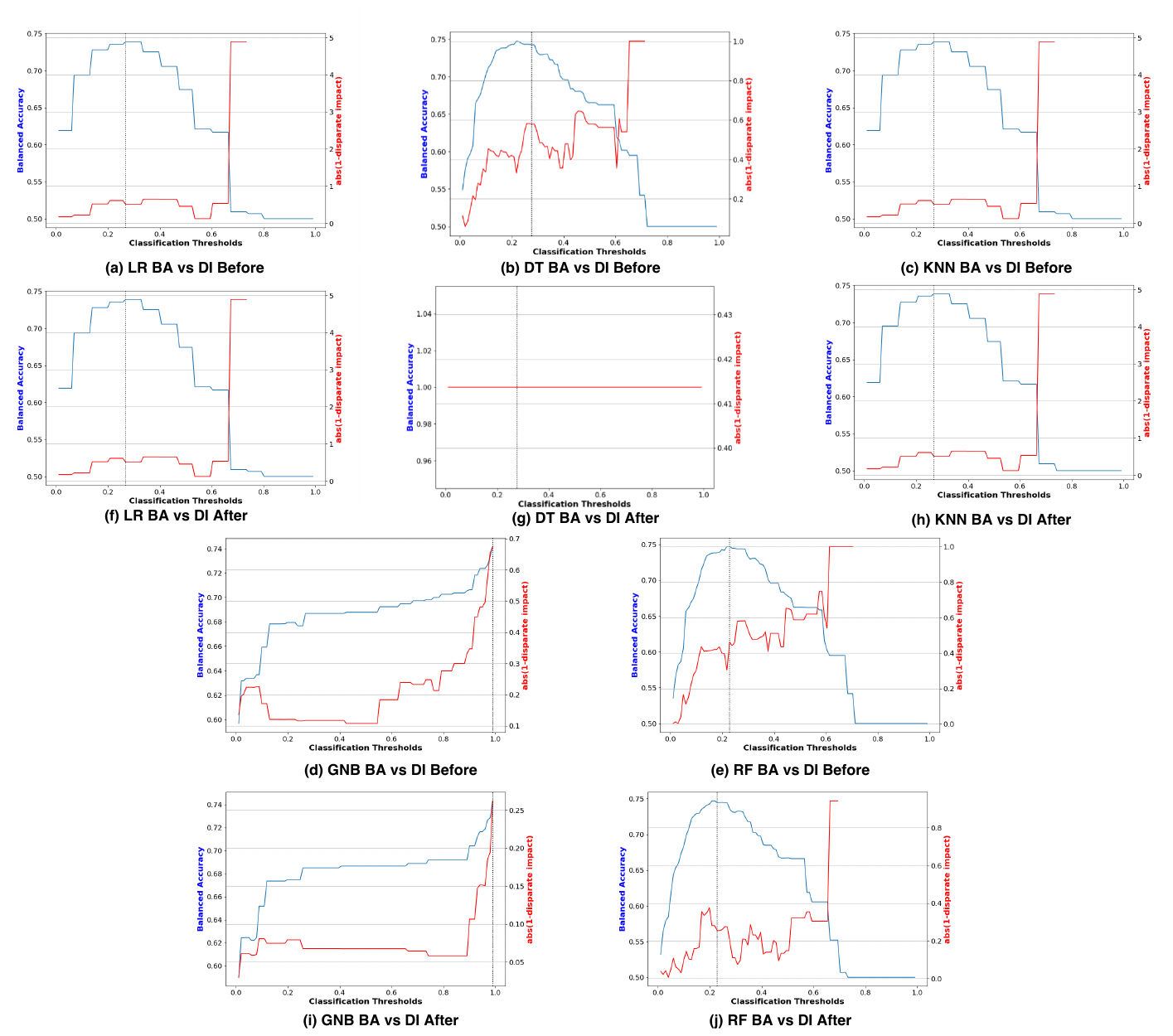}
	\caption{Performance comparison via BA vs DI before and after reweighting samples on Adult income dataset with respect to the protected attribute $Race$.}
	\label{fig:Adult_Race_DI}
\end{figure*}

Figures \ref{fig:Adult_Race_AOD} and \ref{fig:Adult_Race_DI} illustrate the performance comparison before and after reweighting samples with respect to the protected attribute $Race$. Prior to reweighting, various machine learning (ML) models exhibit distinct biases in AOD and DI curves. Following reweighting, biases in ML models are mitigated to varying degrees. Notably, decision tree (DT) models demonstrate bias-free behavior in terms of AOD and DI values. Conversely, biases in logistic regression (LR), k-nearest neighbors (KNN), and Gaussian Naive Bayes (GNB) appear unchanged, while the bias in random forest (RF) becomes more unstable. This suggests that the effectiveness of reweighting samples depends on the specific ML model and may not universally apply.

Table \ref{tab_adult_race} provides a systematic comparison using additional fairness metrics. DT's bias is eliminated in terms of AOD, EOD, and TI values. However, regarding SPD and DI values, bias persists towards the unprivileged group, emphasizing the need for a comprehensive examination of bias. Furthermore, although other ML models experience slight reductions in balanced accuracy (BAs), their fairness is marginally improved, emphasizing the inherent trade-off between BA and fairness.

\begin{figure*}[h!]
	\center
	\includegraphics[width=1.0\linewidth]{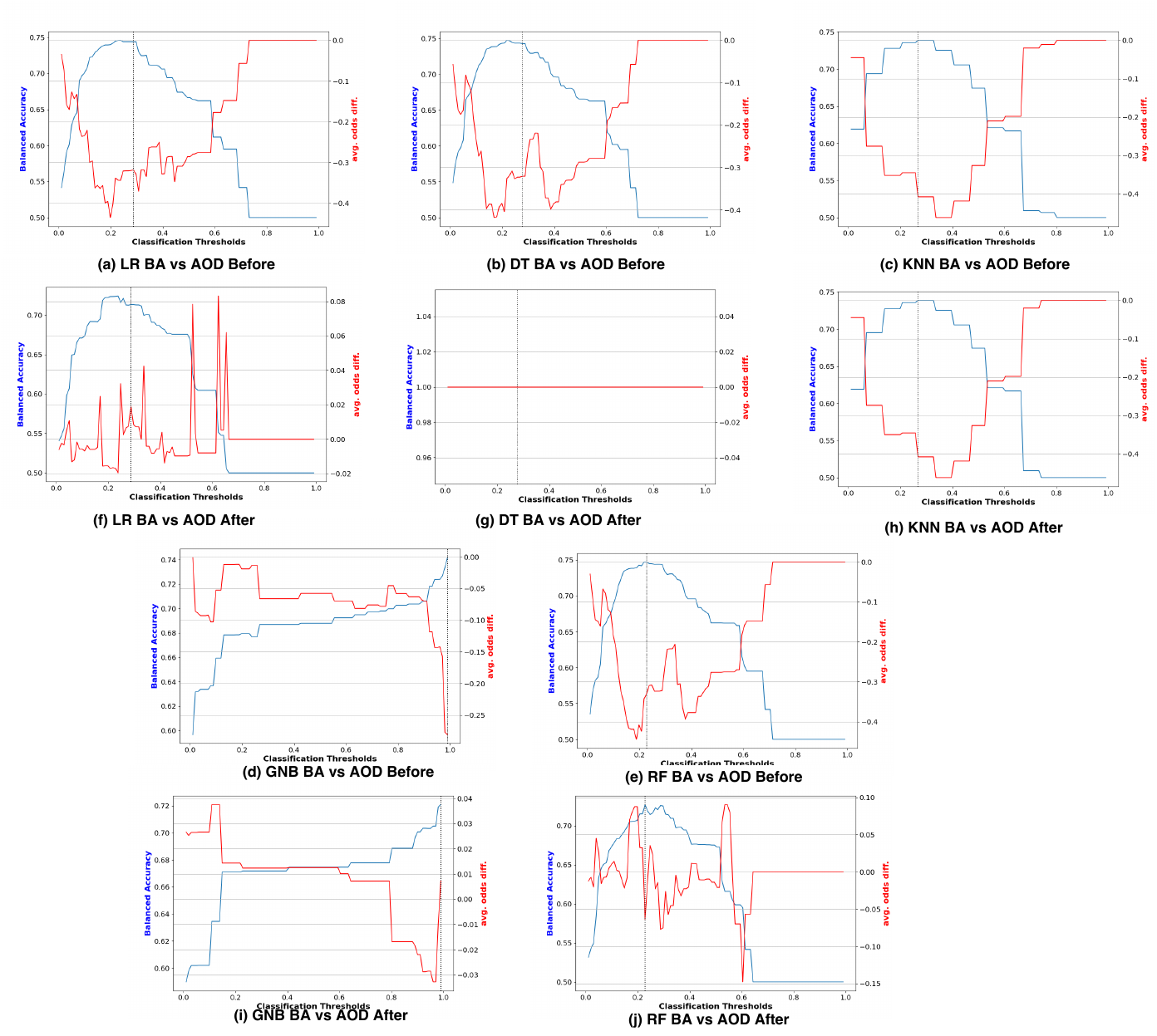}
	\caption{Performance comparison via BA vs AOD before and after reweighting samples on Adult income dataset with respect to the protected attribute $Sex$.}
	\label{fig:Adult_Sex_AOD}
\end{figure*}

\begin{figure*}[h!]
	\center
	\includegraphics[width=1.0\linewidth]{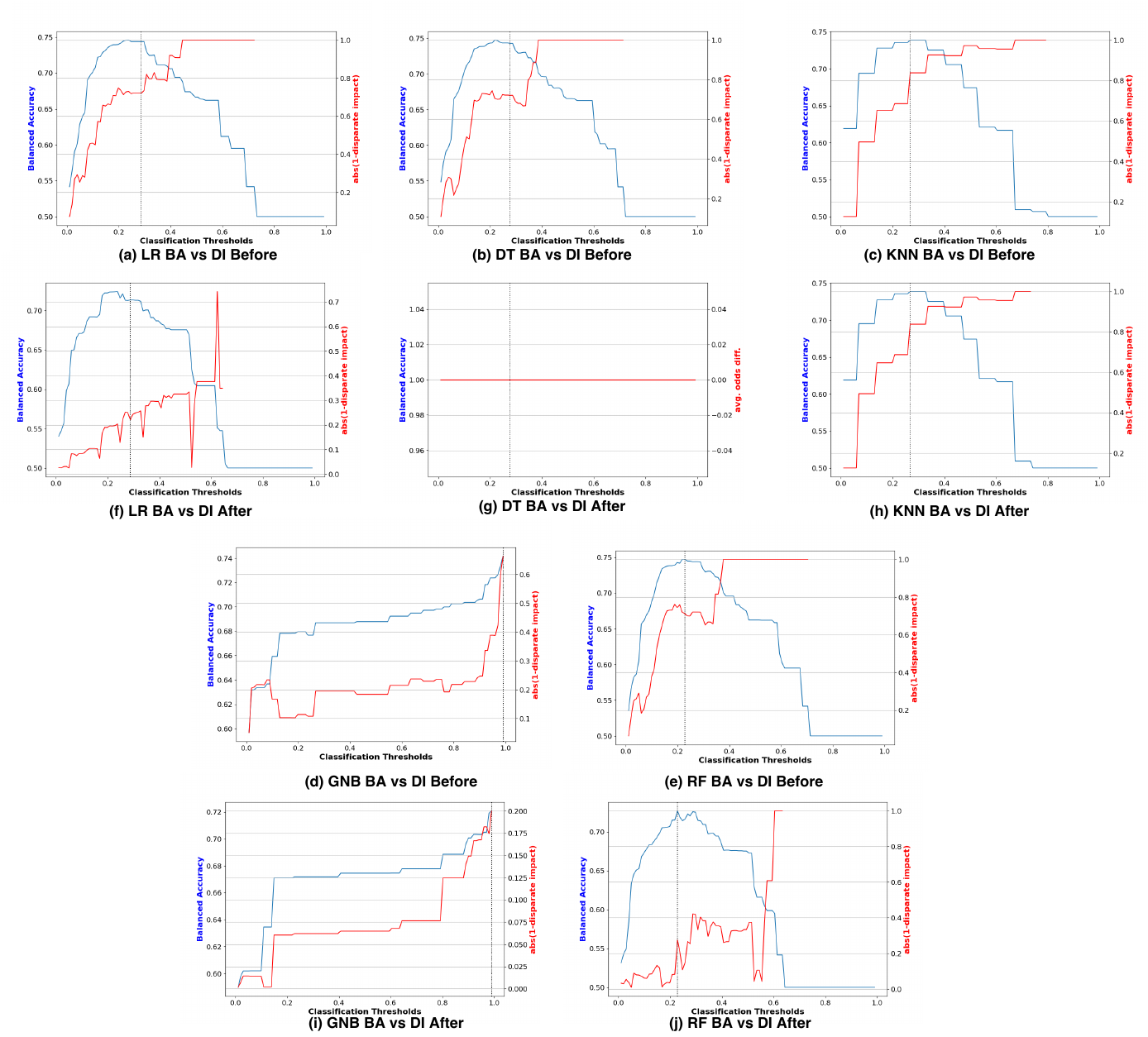}
	\caption{Performance comparison via BA vs DI before and after reweighting samples on Adult income dataset with respect to the protected attribute $Sex$.}
	\label{fig:Adult_Sex_DI}
\end{figure*}

Additionally, Figures \ref{fig:Adult_Sex_AOD} and \ref{fig:Adult_Sex_DI} depict comparison results for the protected attribute $Sex$. Similar observations are noted, where reweighting samples are effective mainly for DT, less so for KNN, and have varying impacts on LR, GNB, and RF. Table \ref{tab_adult_sex} echoes these trends, revealing slight reductions in BA and modest improvements in fairness for other ML models across various fairness metrics (SPD, AOD, DI, EOD, and TI). This underscores the nuanced effectiveness of the same debiasing technique across different ML models.

\begin{table*}[ht]
	\caption{Performance comparison between before and after reweighting samples through one classification metric BA and fairness metrics including SPD, AOD, DI, EOD and TI on Adult income dataset regarding the protected attribute $Race$. } 
        \begin{center}
                \begin{tabular}{|l|cccccc|}
                \hline \multicolumn{7}{|c|}{Performance before reweighting samples}\\ \hline
                    \hline \textbf{Model}  & BA & SPD & AOD & DI & EOD & TI \\ \hline
                    	   Decision Tree					&  0.7426		&  -0.2416		&  -0.1959		&  0.4196 		&  -0.2026		&  0.1130	 \\
                            Gaussian Naïve Bayes			&  0.7416		&  -0.2952		&  -0.2623		&  0.3252 		&  -0.2872		&  0.1111	\\
                            K Nearest Neighbor			&  0.7390		&  -0.1904		&  -0.1409		&  0.4882 		&  -0.1416		&  -0.1416  \\
                            Logistic Regression			&  0.7390		&  -0.1904		&  -0.1409		&  0.4882 		&  -0.1416		&  0.1207	\\
                            Random Forest				&  0.7471		&  -0.2014		&  -0.1336		&  0.5380		&  -0.1097		&  0.1207	\\ \hline
                         
                     \hline
                     \hline \multicolumn{7}{|c|}{Performance after reweighting samples}\\ \hline
                     	   Decision Tree				&  1.0		&  -0.1066		&  0.0		&  0.5863 		&  0.0		&  0.0	 \\
                            Gaussian Naïve Bayes		&  0.7432		&  -0.1147		&  -0.0252		&  0.7379 		&  0.0310		&  0.1058	\\
                            K Nearest Neighbor		&  0.7390		&  -0.1904		&  -0.1409       &  0.4882		&  -0.1416		&  -0.1207  \\
                            Logistic Regression		&  0.7390		&  -0.1904		&  -0.1409		&  0.4882 		&  -0.1416		&  0.1207	\\
                            Random Forest			&  0.7447		&  -0.1072		&  -0.0201		&  0.7449		&  0.0321		&  0.1081	\\ \hline
                           
                            \hline
                      
                \end{tabular}
       \end{center}
         \label{tab_adult_race}
\end{table*}

\begin{table*}[ht]
	\caption{Performance comparison between before and after reweighting samples through one classification metric BA and fairness metrics including SPD, AOD, DI, EOD and TI on Adult income dataset regarding the protected attribute $Sex$. } 
        \begin{center}
                \begin{tabular}{|l|cccccc|}
                \hline \multicolumn{7}{|c|}{Performance before reweighting samples}\\ \hline
                    \hline \textbf{Model}  & BA & SPD & AOD & DI & EOD & TI \\ \hline
                    	   Decision Tree					&  0.7426		&  -0.3608		&  -0.3204		&  0.2785 		&  -0.3775		&  0.1130	 \\
                            Gaussian Naïve Bayes			&  0.7416		&  -0.3353		&  -0.2805		&  0.3369 		&  -0.3184		&  0.1111	\\
                            K Nearest Neighbor			&  0.7390		&  -0.3983		&  -0.4075		&  0.1616 		&  -0.5311		&  -0.1207  \\
                            Logistic Regression			&  0.7437		&  -0.3580		&  -0.3181		&  0.2794 		&  -0.3769		&  0.1129	\\
                            Random Forest				&  0.7471		&  -0.3777		&  -0.3292		&  0.2884		&  -0.3763		&  0.1066	\\ \hline
                         
                     \hline
                     \hline \multicolumn{7}{|c|}{Performance after reweighting samples}\\ \hline
                     	   Decision Tree				&  1.0		&  -0.1910		&  0.0		&  0.3740 		&  0.0		&  0.0	 \\
                            Gaussian Naïve Bayes		&  0.7209		&  -0.0861		&  0.0073		&  0.7997 		&  0.0203		&  0.1192	\\
                            K Nearest Neighbor		&  0.7390		&  -0.3983		&  -0.4075       &  0.1616		&  -0.5311		&  0.1207  \\
                            Logistic Regression		&  0.7134		&  -0.0705		&  0.0188		&  0.7785 		&  0.0293		&  0.1401	\\
                            Random Forest			&  0.7271		&  -0.1386		&  -0.0638		&  0.7220		&  -0.0774		&  0.1065	\\ \hline
                         
                     \hline
                \end{tabular}
       \end{center}
         \label{tab_adult_sex}
\end{table*}

\subsubsection{COMPAS}

\begin{figure*}[h!]
	\center
	\includegraphics[width=1.0\linewidth]{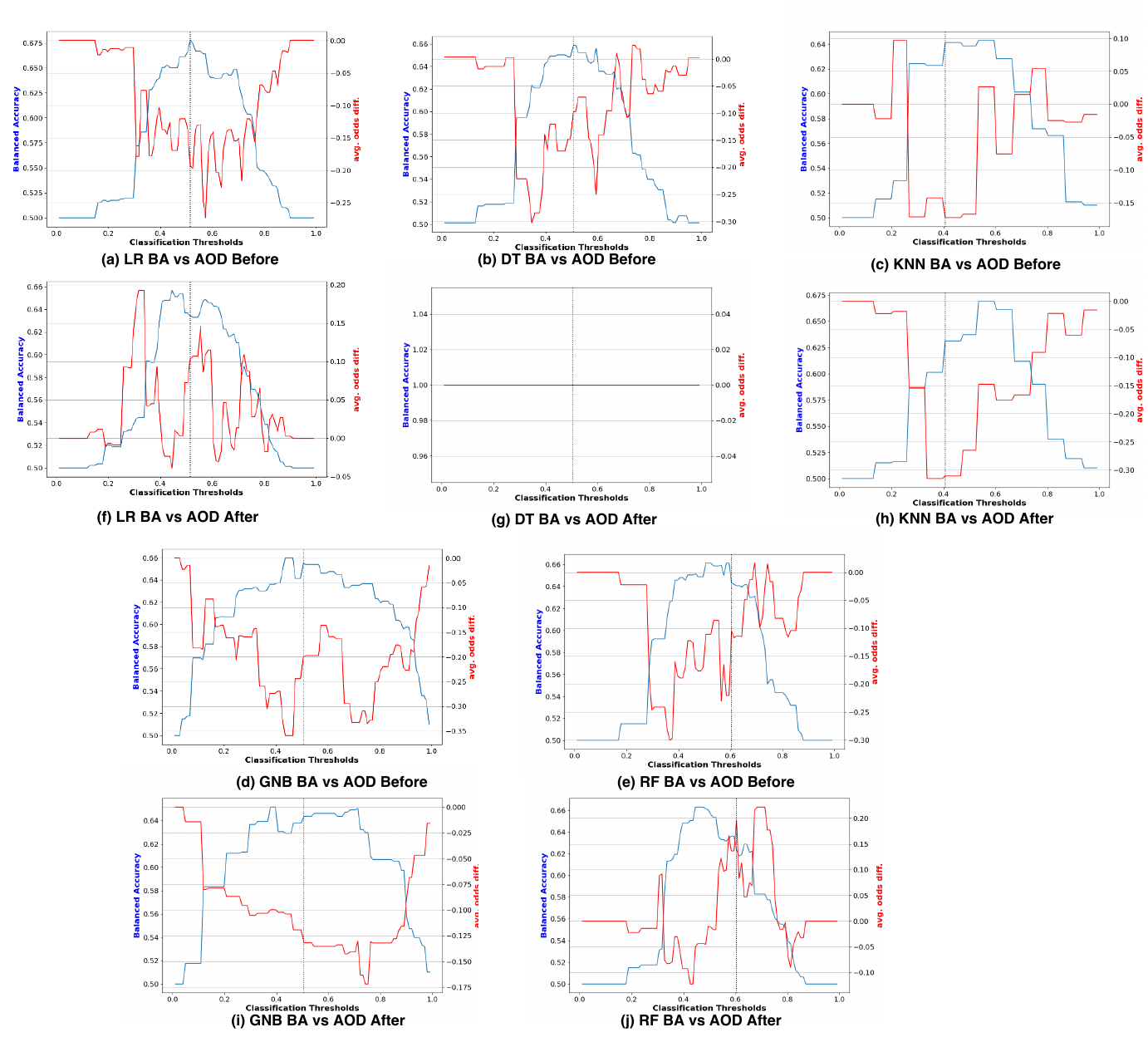}
	\caption{Performance comparison via BA vs AOD before and after reweighting samples on COMPAS dataset with respect to the protected attribute $Race$.}
	\label{fig:Compas_Race_AOD}
\end{figure*}

\begin{figure*}[h!]
	\center
	\includegraphics[width=1.0\linewidth]{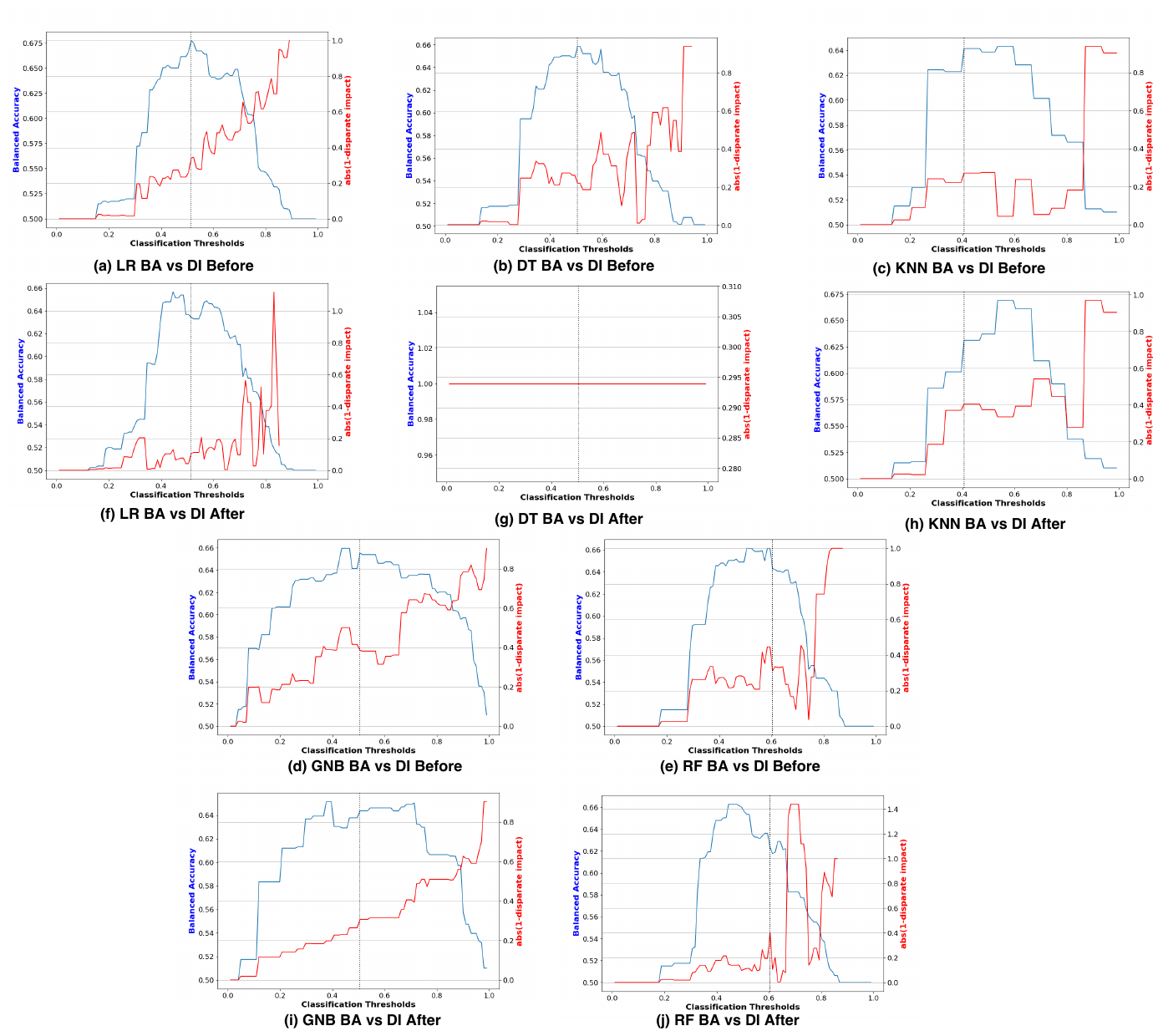}
	\caption{Performance comparison via BA vs DI before and after reweighting samples on COMPAS dataset with respect to the protected attribute $Race$.}
	\label{fig:Compas_Race_DI}
\end{figure*}

To thoroughly assess the effectiveness of reweighting samples in mitigating bias, Figures \ref{fig:Compas_Race_AOD} and \ref{fig:Compas_Race_DI} provide a performance comparison before and after reweighting samples for the protected attribute $Race$ in the COMPAS dataset. Similarly, prior to reweighting, various machine learning (ML) models exhibit diverse biases in AOD and DI curves. Post-reweighting, DT models demonstrate a lack of bias in AOD and DI values. Conversely, biases in LR, KNN, GNB, and RF appear unchanged, as seen in the curves of AOD and DI.

Table \ref{tab_compas_race} offers a systematic comparison using additional fairness metrics on the COMPAS dataset. Similar to the Adult dataset, DT's bias is eradicated concerning AOD, EOD, and TI values. However, biases persist in terms of SPD and DI values, showing a continued bias toward the unprivileged group. Additionally, while other ML models experience slight reductions in balanced accuracy (BA), their fairness is marginally improved.

\begin{figure*}[h!]
	\center
	\includegraphics[width=1.0\linewidth]{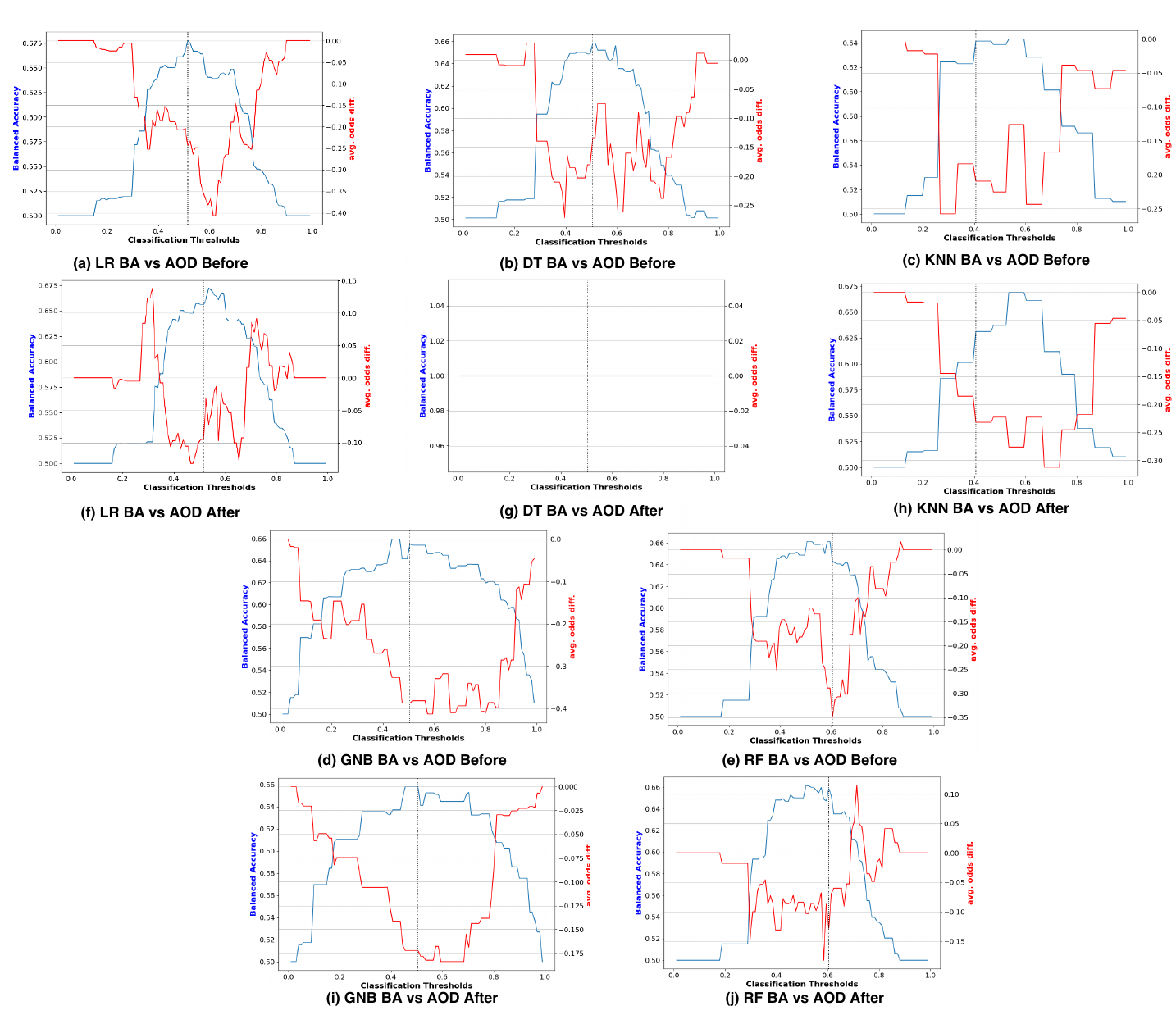}
	\caption{Performance comparison via BA vs AOD before and after reweighting samples on COMPAS dataset with respect to the protected attribute $Sex$.}
	\label{fig:Compas_Sex_AOD}
\end{figure*}

\begin{figure*}[h!]
	\center
	\includegraphics[width=1.0\linewidth]{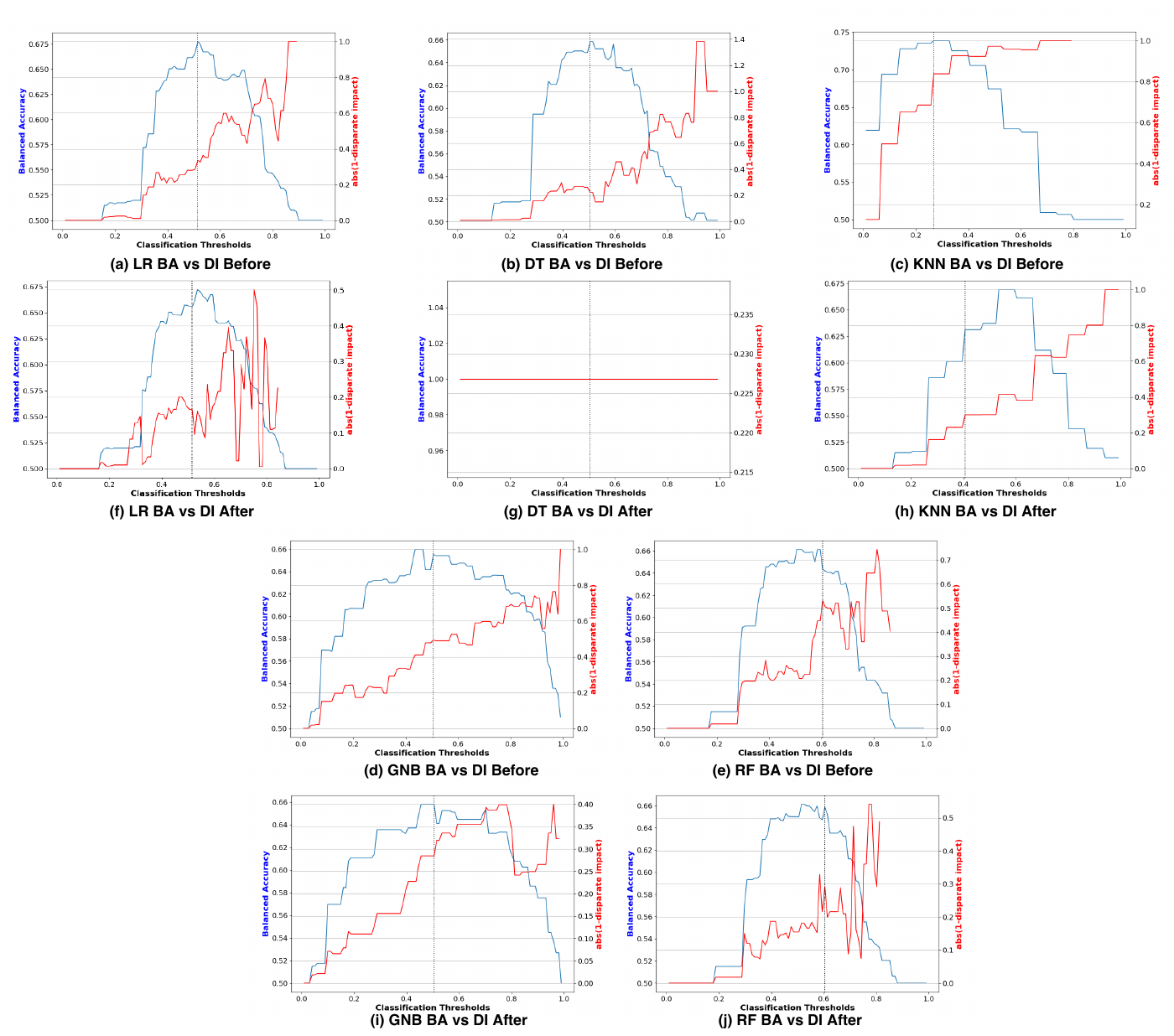}
	\caption{Performance comparison via BA vs DI before and after reweighting samples on COMPAS dataset with respect to the protected attribute $Sex$.}
	\label{fig:Compas_Sex_DI}
\end{figure*}

Moreover, Figures \ref{fig:Compas_Sex_AOD} and \ref{fig:Compas_Sex_DI} present comparison results for the protected attribute $Sex$. Similar observations are made, where reweighting samples are effective primarily for DT but less so for KNN, GNB, and RF. Notably, LR exhibits more significant effects after reweighting. Table \ref{tab_compas_sex} reveals similar trends, with slight reductions in BA and modest improvements in fairness across various metrics (SPD, AOD, DI, EOD, and TI).

In summary, the figures and tables illustrate the impact of reweighting samples on bias mitigation in traditional machine learning models for both the Adult income and COMPAS datasets. Notably, decision tree (DT) models showcase effective bias reduction, while logistic regression (LR), k-nearest neighbors (KNN), and Gaussian Naive Bayes (GNB) exhibit more resistance to debiasing. The trade-off between balanced accuracy (BA) and fairness is evident. Results differ across protected attributes, emphasizing the nuanced effectiveness of reweighting. Comprehensive assessments, including additional fairness metrics, reveal the complexity of bias dynamics. This study highlights \emph{the model-specific nature of reweighting sample effectiveness in achieving fairness in traditional machine learning models}.

\begin{table*}[ht]
	\caption{Performance comparison between before and after reweighting samples through one classification metric BA and fairness metrics including SPD, AOD, DI, EOD and TI on COMPAS dataset regarding the protected attribute $Race$. } 
        \begin{center}
                \begin{tabular}{|l|cccccc|}
                \hline \multicolumn{7}{|c|}{Performance before reweighting samples}\\ \hline
                    \hline \textbf{Model}  & BA & SPD & AOD & DI & EOD & TI \\ \hline
                    	   Decision Tree					&  0.6586		&  -0.1516		&  -0.0970		&  0.7791		&  -0.1212		&  0.1835	 \\
                            Gaussian Naïve Bayes			&  0.6553		&  -0.2483		&  -0.1994		&  0.6155 		&  -0.1980		&  0.2382	\\
                            K Nearest Neighbor			&  0.6414		&  -0.2139		&  -0.1727		&  0.7282 		&  -0.1131		&  -0.1607  \\
                            Logistic Regression			&  0.6774		&  -0.2494		&  -0.1927		&  0.6600 		&  -0.1877		&  0.1774	\\
                            Random Forest				&  0.6432		&  -0.1539		&  -0.1057		&  0.6873		&  -0.1166		&  0.3003	\\ \hline
                         
                     \hline
                     \hline \multicolumn{7}{|c|}{Performance after reweighting samples}\\ \hline
                     Decision Tree		&  1.0		&  -0.1769		&  0.0		&  0.7060 		&  0.0		&  0.0	 \\
                            Gaussian Naïve Bayes		&  0.6437		&  -0.1782		&  -0.1318		&  0.6926 		&  -0.1251		&  0.2594	\\
                            K Nearest Neighbor		&  0.6311		&  -0.3432		&  -0.3105       &  0.5945		&  -0.2391		&  -0.1762  \\
                            Logistic Regression		&  0.6342		&  0.0546		&  0.1042		&  1.1062 		&  0.1215		&  0.2257	\\
                            Random Forest			&  0.6234		&  0.1459		&  0.1953		&  1.4012		&  0.1977		&  0.2867	\\ \hline
                         
                     \hline

                \end{tabular}
       \end{center}
         \label{tab_compas_race}
\end{table*}

\begin{table*}[ht]
	\caption{Performance comparison between before and after reweighting samples through one classification metric BA and fairness metrics including SPD, AOD, DI, EOD and TI on COMPAS dataset regarding the protected attribute $Sex$. } 
        \begin{center}
                \begin{tabular}{|l|cccccc|}
                \hline \multicolumn{7}{|c|}{Performance before reweighting samples}\\ \hline
                    \hline \textbf{Model}  & BA & SPD & AOD & DI & EOD & TI \\ \hline
                    	   Decision Tree				&  0.6586		&  -0.1637		&  -0.1340		&  0.7759 		&  -0.0597		&  0.1835	 \\
                            Gaussian Naïve Bayes		&  0.6553		&  -0.4129		&  -0.3877		&  0.5066 		&  -0.3090		&  0.2382	\\
                            K Nearest Neighbor		&  0.6414		&  -0.2336		&  -0.2095		&  0.7256 		&  -0.1350		&  0.1607  \\
                            Logistic Regression		&  0.6774		&  -0.2724		&  -0.2439		&  0.6631 		&  -0.1392		&  0.1774	\\
                            Random Forest			&  0.6432		&  -0.3759		&  -0.3484		&  0.4700		&  -0.3002		&  0.3003	\\ \hline
                            
                                 \hline
                         \hline \multicolumn{7}{|c|}{Performance after reweighting samples}\\ \hline
                         	   Decision Tree				&  1.0		&  -0.1383		&  0.0		&  0.7732 		&  0.0		&  0.0	 \\
                            Gaussian Naïve Bayes		&  0.6581		&  -0.1998		&  -0.1720		&  0.7154 		&  -0.0899		&  0.2146	\\
                            K Nearest Neighbor		&  0.6311		&  -0.2551		&  -0.2318         &  0.7003		&  -0.1708		&  0.1762 \\
                            Logistic Regression		&  0.6562		&  -0.1188		&  -0.0946		&  0.8342 		&  0.0111		&  0.1730	\\
                            Random Forest			&  0.6585		&  -0.1615		&  -0.1279		&  0.7081		&  -0.0760		&  0.2776	\\ \hline

                     \hline
                     
                \end{tabular}
       \end{center}
         \label{tab_compas_sex}
\end{table*}

\section{Related Work }
\label{sec5}

The fairness of machine learning (ML) models has become one of the most pivotal challenges of the decade~\cite{shaham2023holistic}. Intended to intelligently prevent errors and biases in decision-making, ML models sometimes unintentionally become sources of bias and discrimination in society. Concerns have been raised about various forms of unfairness in ML, including racial biases in criminal justice systems, disparities in employment, and biases in loan approval processes~\cite{angwin2022machine}. The entire life cycle of an ML model, covering input data, modeling, evaluation, and feedback, is susceptible to both external and inherent biases, potentially resulting in unjust outcomes.

Techniques for mitigating bias in machine learning models can be categorized into pre-processing, in-processing, and post-processing methods~\cite{caton2020fairness}. Pre-processing acknowledges that data itself often introduces bias, with distributions of sensitive or protected variables being discriminatory or imbalanced. This approach modifies sample distributions or transforms data to eliminate discrimination during training~\cite{kamiran2012data}. It is considered the most flexible part of the data science pipeline, making no assumptions about subsequent modeling techniques~\cite{du2018data}. In-processing adjusts modeling techniques to counter biases and incorporates fairness metrics into model optimization~\cite{jiang2020identifying}. Post-processing addresses unfair model outputs, applying transformations to enhance prediction fairness~\cite{kim2019multiaccuracy}.

Pre-processing assumes that the disparate impact of the trained classifier mirrors that of the training data. Techniques include massaging the dataset by adjusting mislabeled class labels due to bias~\cite{feldman2015certifying, kamiran2009classifying} and reweighting training samples to assign greater importance to sensitive ones~\cite{doherty2012information, kamiran2012data}.

\section{Conclusion and Future Work}
\label{sec7}

This study conducted thorough validation on the application of reweighting samples to address bias in binary classification using traditional machine learning models. It provides a systematic insight into the effectiveness of various traditional classification algorithms concerning different protected attributes, contributing to the advancement of fairness-enhanced AI. Future research will extend this exploration by incorporating more advanced machine learning algorithms, including generative AI models. Additionally, the study will incorporate new datasets and explore a broader range of protected attributes to further enhance the validation process.

\section*{Acknowledgment}
\label{acknowledgement}
This research work is supported by NSF  under award number 2323419. Any opinions, findings, and conclusions or recommendations expressed in this work are those of the
author(s) and do not necessarily reflect the views of NSF.

\ifCLASSOPTIONcaptionsoff
  \newpage
\fi


\bibliographystyle{IEEEtran}
\bibliography{Reference}

\end{document}